\pdfoutput=1

\documentclass[11pt]{article}

\usepackage[final]{acl}

\usepackage{times}
\usepackage{latexsym}
\usepackage{float}
\usepackage{xcolor}
\usepackage{tcolorbox}
\usepackage[T1]{fontenc}

\usepackage[utf8]{inputenc}

\usepackage{microtype}

\usepackage{inconsolata}

\usepackage{graphicx}

\title{Hybrid Annotation for Propaganda Detection: Integrating LLM Pre-Annotations with Human Intelligence}

\author{
\textbf{Ariana Sahitaj}\textsuperscript{1,2}\thanks{\ Equal contribution} \quad
\textbf{Premtim Sahitaj}\textsuperscript{1,2}\footnotemark[1] \quad
\textbf{Veronika Solopova}\textsuperscript{1,2} \\
\textbf{Jiaao Li}\textsuperscript{1,2} \quad
\textbf{Sebastian Möller}\textsuperscript{1,2} \quad
\textbf{Vera Schmitt}\textsuperscript{1,2} \\
\textsuperscript{1}Quality and Usability Lab, Technische Universität Berlin, Germany \\
\textsuperscript{2}German Research Center for Artificial Intelligence (DFKI), Berlin, Germany \\
{\small \texttt{ariana.sahitaj@campus.tu-berlin.de}}
}

\usepackage{url}
\begin{document}
\maketitle
\begin{abstract}

Propaganda detection on social media remains challenging due to task complexity and limited high-quality labeled data. This paper introduces a novel framework that combines human expertise with Large Language Model (LLM) assistance to improve both annotation consistency and scalability. We propose a hierarchical taxonomy that organizes 14 fine-grained propaganda techniques \cite{martino2020semeval} into three broader categories, conduct a human annotation study on the HQP dataset \cite{maarouf2023hqp} that reveals low inter-annotator agreement for fine-grained labels, and implement an LLM-assisted pre-annotation pipeline that extracts propagandistic spans, generates concise explanations, and assigns local labels as well as a global label. A secondary human verification study shows significant improvements in both agreement and time-efficiency. Building on this, we fine-tune smaller language models (SLMs) to perform structured annotation. Instead of fine-tuning on human annotations, we train on high-quality LLM-generated data, allowing a large model to produce these annotations and a smaller model to learn to generate them via knowledge distillation. Our work contributes towards the development of scalable and robust propaganda detection systems, supporting the idea of transparent and accountable media ecosystems in line with SDG 16. The code is publicly available at our GitHub repository\footnote{\url{https://github.com/XplaiNLP/NLP4PI_2025_submission}}.\\
\\
    \textbf{Content Warning:} This paper contains examples of Russian propaganda, some of which contain misleading, or offensive claims. These are provided for academic analysis and do not reflect the authors' views.
\end{abstract}

\section{Introduction}

Fake news and disinformation have become a significant challenge, particularly in geopolitical conflicts like the Russia-Ukraine war \cite{perez2022strategic}. Disinformation campaigns strategically manipulate public opinion and shape narratives \cite{wardle2017information, zhdanova2017computational}, with pro-Russian biases linked to reduced ability to identify propaganda \cite{erlich2023pro}. Propaganda, defined as \textit{"the deliberate and systematic attempt to shape perceptions, manipulate cognitions, and direct behavior to achieve a response that furthers the desired intent of the propagandist"} \cite{lock2020organizational, jowett2018propaganda}, lies at the core of these campaigns. Detecting such manipulative content is critical for preserving public trust and safeguarding democratic processes \cite{bayer2021disinformation}.
\begin{figure}
    \centering
    \includegraphics[width=1\linewidth]{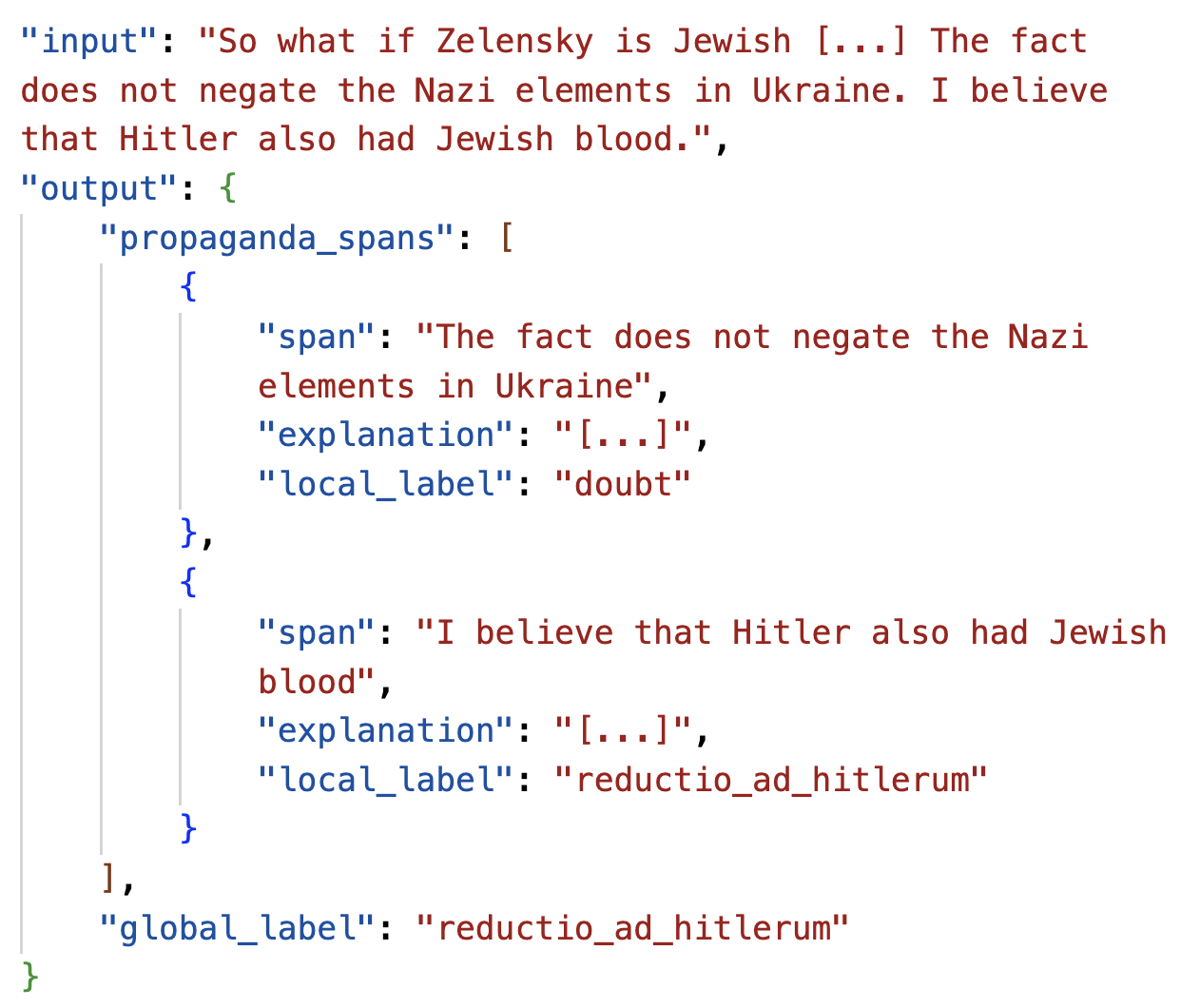}
    \caption{Our proposed LLM output for a reduced input tweet from the HQP dataset \cite{maarouf2023hqp} where it was initially weak-labeled as "slogans".}
    \label{fig:example}
\end{figure}
While propaganda in long-form text is well studied \cite{martino2020semeval}, short-form propaganda remains more challenging due to limited annotated data, sparse context, and the use of informal language, abbreviations, and hashtags \cite{vijayaraghavan2022tweetspin}. Although automated methods for disinformation and propaganda detection have advanced \cite{plikynas2025systematic}, the task remains difficult. Subtle linguistic cues, context-dependent interpretations, and low inter-annotator agreement highlight the complexity of human annotations \cite{hasanain2023large, srba2024survey}, particularly in fine-grained classification \cite{hasanain-etal-2024-gpt, martino2020semeval}, as propaganda often exploits cognitive biases and undermines critical thinking, making individuals more susceptible to conspiratorial narratives \cite{tanvir2024information, sahitaj2024towards}.
Propaganda detection aligns with the United Nations Sustainable Development Goal (SDG) 16\footnote{https://sdgs.un.org/goals/goal16}, which promotes peaceful, inclusive societies and effective institutions. Misinformation and propaganda undermine these aspirations by fueling social divisions, eroding trust in institutions, and obstructing transparent communication \cite{mwangi2023technology}, especially when amplified by automated bots \cite{zhdanova2017computational}.\\
In this work, we propose a methodology that advances propaganda detection through the following five key contributions: First, we develop a fine-grained propaganda taxonomy that categorizes 14 distinct techniques by \citet{martino2020semeval} into three broader groups based on their intent: those that trigger emotional responses, those that simplify or distort complex issues, and those that undermine trust through authority and group dynamics. Second, we conduct an initial human annotation study on a statistically significant subset of propagandistic tweets from the HQP dataset \cite{maarouf2023hqp}. This study highlights the challenges of manual fine-grained labeling, revealing that the process is highly subjective, time-consuming, and prone to low inter-annotator agreement. Third, to overcome these limitations, we propose a novel LLM-assisted annotation methodology. In our pipeline, LLMs first extract relevant propaganda spans from the text, explain why these spans are considered propagandistic, and then assign fine-grained labels at the span level before determining a global label for the entire post. Fourth, we perform a secondary human verification study on a stratified sample of LLM-annotated posts. In this stage, human annotators are presented with the extracted spans and their local labels, and tasked with annotating the global propaganda label. We observe that annotation agreement increases, and time investment is reduced by introducing LLMs as pre-annotation tool. Finally, we fine-tune small language models on the LLM-generated annotations to perform structured span-based labeling and explanation, enabling scalable training through knowledge distillation without relying on human-labeled data.


\begin{figure*}
    \centering
    \includegraphics[width=1\linewidth]{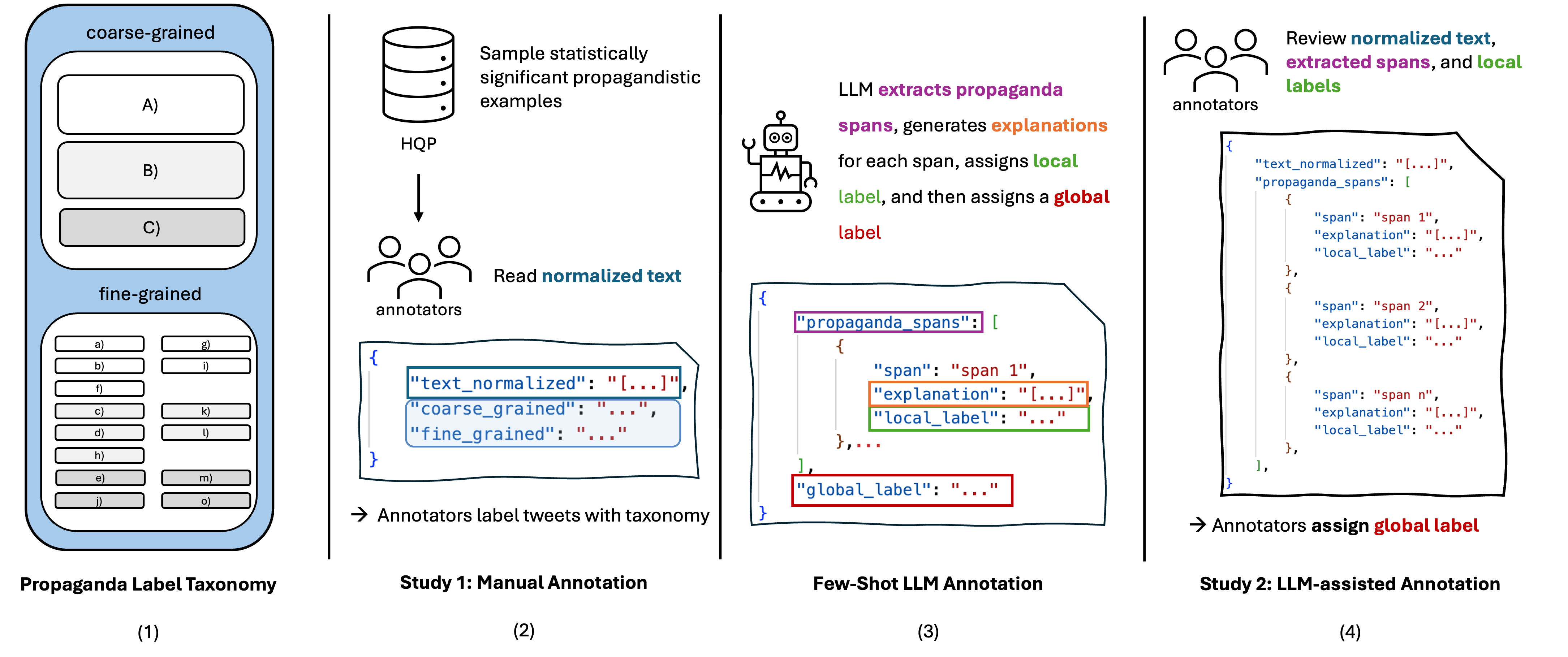}
    \caption{Methodological Overview}
    \label{fig:overview}
\end{figure*}

\section{Related Work}

Early research on automatic propaganda detection approached the problem at the document level, aiming to classify entire news articles \cite{rashkin2017truth}. For instance, some systems labeled texts into four broad categories (trusted, satire, hoax, or propaganda) \cite{rashkin2017truth}, while others framed it as a binary task (propaganda, non-propaganda) \cite{barron2019proppy}, which limited granularity and explainability \cite{martino2019fine}. An advance came with the work of \citet{martino2019fine}, who introduced span-level analysis with the PTC corpus, which comprises news articles annotated at the sentence level and fragment level with 18 distinct propaganda techniques. This scheme was adopted by the SemEvak-2020 Shared Task \cite{martino2020semeval} which consolidated the 18 techniques into a set of 14 widely used labels \cite{martino2020semeval, sprenkamp2023large, abdullah2022detecting}, that we also follow in our work. Early models used BERT-based architectures to perform span identification and technique classification \cite{da2019findings}. Building on this, recent work explores how LLMs can further enhance propaganda detection, in terms of reducing annotation time and cost while improving label agreement and quality across classification tasks \cite{alizadeh2025open, gilardi2023chatgpt, ding2022gpt}. However, the use of LLMs may also exhibit stronger systematic bias than human annotators, especially in politically sensitive contexts \cite{vera2024bias}, and may suffer from generation-related issues such as hallucinations \cite{lee2023mathematical}. Within propaganda detection, \citet{jose2025large} evaluated GPT-3.5, GPT-4, and Claude on identifying six propaganda techniques in news articles. \citet{hasanain2023large} employed GPT-4 as an LLM-as-Annotator approach to annotate Arabic text spans with 23 propaganda techniques using multilabel and sequence tagging tasks, and trained BERT-based models on the generated annotations. Similarly, \citet{sprenkamp2023large} examined the performance of multiple GPT-3 and GPT-4 variants for multi-label classification of 14 propaganda techniques at article-level using the SemEval-2020 Task 11 dataset \cite{martino2020semeval}, employing a range of prompt engineering and fine-tuning strategies. Their results show that GPT-4 can approach state-of-the-art performance. Our work builds on these efforts by grouping the 14 fine-grained techniques \cite{martino2020semeval} into a novel coarse-grained taxonomy of three broader categories to support human annotator clarity and enable hierarchical modeling. By using a fully open-source LLM (LLaMA3-70B), we extract propaganda spans from tweets and assign fine-grained local labels based on the 14 techniques from \citet{martino2020semeval}. In addition, it assigns a global propaganda label that captures the tweet’s overall framing. While the LLM also generates explanations for why each span was classified as propagandistic, these are not shown to human annotators but are used as an intermediate reasoning step to guide models towards their prediction. Moreover, we distill four small student models on the generated outputs of the larger model as teacher to enable propaganda span in resource-constrained environments through an open-source modeling pipeline.

\section{Methodology and Results}
In this Section, we outline our novel methodology that combines human expertise with computational techniques, as displayed in Figure \ref{fig:overview}, and their results. We first define a labeling framework for both coarse-grained and fine-grained categories in Section \ref{sec:propaganda_labels}. Next, we describe our human annotation study (Study 1, see Section \ref{human annotation study 1}) on the HQP dataset \cite{maarouf2023hqp}. We then detail our LLM few-shot inference in Section \ref{few-shot llm annotation} and annotation approach, to automatically extract propaganda spans, generate explanations, and assign fine-grained labels, followed by a second human verification study (Study 2, see Section \ref{human annotation study 2}). Finally, we fine-tune SMLs via knowledge distillation in Section \ref{knowledge distillation}.

\subsection{Propaganda Label Taxonomy}
\label{sec:propaganda_labels}

Annotating text for propaganda techniques is a highly complex task, as it is influenced by subjectivity, cognitive biases, personal experiences and the subtle variations in meaning that arise from different cultural and linguistic contexts \cite{sprenkamp2023large}. Prior work has highlighted that distinguishing between multiple fine-grained techniques can be particularly demanding, leading to low inter-annotator agreement and making it difficult to maintain consistency across annotations. \cite{hasanain-etal-2024-gpt} \\
To investigate this problem, we survey the literature and aggregate definitions from previous works, most notably the 14 propaganda techniques introduced by \citet{martino2020semeval}, which refined an earlier set of 18 techniques proposed by \citet{martino2019fine} and later applied by \citet{sprenkamp2023large} and \citet{abdullah2022detecting} to analyze and label propaganda techniques in text. \\
In our framework, the fine-grained propaganda techniques are organized in broader, coarse-grained categories according to their manipulative intent and rhetorical function. Detailed definitions of the techniques can be found in the Appendix \ref{sec:fine-grained labels}. This hierarchical framework aims to reduce cognitive load for annotators and improve labeling consistency by first categorizing propaganda into conceptual groups before applying fine-grained classifications. It also enables us to evaluate the fine-grained predictions within the context of the coarse-grained labeling system in the subsequent analysis. The three coarse-grained categories are as follows:
\begin{itemize}
    \item[(A)] \textbf{Emotional Appeals to Influence Opinions and Behaviors.} Techniques that exploit emotions to influence opinions or actions, often bypassing rational analysis. These methods use emotionally charged language, imagery, or ideas to evoke strong feelings. It includes the following techniques: loaded language, name calling, labeling, appeal to fear/prejudice, flag-waving, slogans.
    \item[(B)]\textbf{Simplification and Distortion Strategies.} Techniques that distort reality by presenting complex issues in oversimplified or misleading ways. These methods often aim to reduce critical thinking and encourage binary or superficial understanding. Here, the following techniques are included: repetition, exaggeration or minimization, causal oversimplification, black-and-white fallacy, thought-terminating clichés.
    \item[(C)] \textbf{Manipulating Trust, Authority, and Rational Discourse.} Techniques that undermine trust, exploit authority, discredit opponents, or manipulate group dynamics to shift opinions. These methods often redirect attention or leverage associations to influence perceptions of credibility or legitimacy. This includes the following techniques: doubt, appeal to authority, whataboutism, straw man, red herring, bandwagon, reductio ad hitlerum.
\end{itemize}

\subsection{Study 1: Human Annotation}
\label{human annotation study 1}

In this initial study, we aim to replicate previous findings from  \citet{hasanain-etal-2024-gpt} that emphasize the challenges of annotating fine-grained propaganda techniques, most notably, the low inter-annotator agreement (IAA) observed in such tasks. For studying the annotation of fine-grained labels, we utilize the HQP dataset \cite{maarouf2023hqp}, which comprises 29,596 tweets annotated for binary propaganda detection within the context of Russian propaganda. Out of these, 4,534 tweets were previously identified as propagandistic.\\
Assuming that the binary classification of propaganda versus non-propaganda is reliable, we confined our analysis to the subset of tweets labeled as propaganda. This focus allowed us to isolate the task of assigning detailed, fine-grained labels without the confounding effects of binary misclassification. Based on a 5\% margin of error at a 95\% confidence level and following established sample size estimation methods \citep{ahmed2024choose}, a sample of $n = 355$ was selected from the 4,534 tweets labeled as propaganda. While this sample is statistically sufficient to estimate proportions, we consider this a pilot study to explore annotation feasibility and qualitative patterns rather than claiming full representativeness of the corpus.

\subsubsection{Setup}
\label{setup1}

Initially, the annotators were provided with the HQP annotation guidelines \cite{maarouf2023hqp}, which define propaganda as \textit{deliberate expressions aimed at influencing opinions}, with a specific focus on Russian propaganda in the context of the Russo-Ukrainian conflict. This ensured a common understanding of the binary classification of tweets as propagandistic. Subsequently, they received a supplementary annotation guideline that included the previously introduced definitions and concrete examples of both coarse-grained and fine-grained propaganda categories. Annotators were instructed to first select the most appropriate coarse-grained category and then assign the single most significant fine-grained label for each tweet.



\subsubsection{Results}
The first human annotation study required three annotators to label each tweet using both the predefined coarse-grained categories and the more detailed fine-grained labels. The coarse-grained labels achieved a moderate level of consensus, as seen in Table \ref{tab:agreement_metrics1}.\\
\begin{table}[h]
  \centering
    \small
  \begin{tabular}{lcc}
    \hline

    \hline
    \textbf{Metric}          & \textbf{Coarse} & \textbf{Fine} \\
    \hline
    Raw Agreement $2/3$      &  0.8845  &   0.4761 \\
    Raw Agreement $3/3$      &  0.2789      &   0.0761 \\
    Krippendorff's Alpha     &   0.2065     &   0.1233 \\
    \hline
  \end{tabular}
    \caption{Inter-Annotator Agreement Metrics for Coarse- and Fine-Grained Propaganda Annotations in Round 1.}
  \label{tab:agreement_metrics1}
\end{table}

Specifically, the raw agreement for coarse-grained annotations reached 88.45\% with a $2/3$ majority but dropped to 27.89\% when full $3/3$ consensus was required. The fine-grained labeling presented greater challenges, with the raw agreement ($2/3$) being 47.61\%, while the full agreement reached only 7.61\%. The corresponding Krippendorff's Alpha values of coarse and fine-grained labels further underscore the limitations in obtaining consistent fine-grained annotations. A more detailed analysis in Table \ref{tab:agreement_x} reveals that fine-grained agreement improves substantially when annotators already agree on the coarse-grained category.

\begin{table}[ht]
  \centering
    \small
  \begin{tabular}{lcc}
    \hline
    \textbf{Subset}          & \textbf{$2/3$ Fine} & \textbf{$3/3$ Fine} \\
    \hline
    $2/3$ Coarse               & 0.4372   & 0.0000 \\
    $3/3$ Coarse               & 0.7475  & 0.2727 \\
    \hline
  \end{tabular}
    \caption{Fine-grained agreement rates conditioned on prior majority $2/3$ or full $3/3$ agreement on coarse labels.}
  \label{tab:agreement_x}
\end{table}


In the guidelines for the annotation of the HQP dataset \cite{maarouf2023hqp}, annotators were asked to label the entire tweet as propagandistic, even if only some segments of the text contain propagandistic content. While we followed this notion for our own annotation of fine-grained labels, our analysis revealed that many tweets comprised multiple segments, each potentially associated with different propaganda labels. This complexity made applying a single definite label to the entire document challenging, as annotators not only had to differentiate among 14 possible labels but also rank the labels based on their impact, so that they could choose the most prominent one. This additional layer of subjectivity and specificity, also contributing to an average annotation time of 151.70 seconds per instance, underscores the need to explore alternative annotation strategies, such as LLM-assisted pre-annotation, as discussed in the following sections.

\subsection{Few-Shot LLM Annotation}
\label{few-shot llm annotation}

Based on the findings of Study 1, we extend the annotation approach by implementing an LLM to extract segments of potential propagandistic content and assign labels at two levels. In this approach, the LLM is tasked with three subtasks: (i) extracting spans from the presented tweet that likely contain propagandistic language, (ii) generating concise explanations for why each span was classified as propagandistic, and (iii) assigning a fine-grained local label to each extracted span as well as a global label for the entire tweet.\\
We employ few-shot inference with llama3.3-70B-Instruct model \cite{llama3modelcard}. Specifically, we create a synthetic few-shot example for each of the fine-grained propaganda labels and incorporate the corresponding label definitions into the system prompt. Each example is manually constructed to reflect a typical use of the respective technique. Three of the authors review each example for clarity and fit. We utilize structured generation to ensure that outputs can be easily parsed and evaluated \cite{willard2023efficient}. No additional background knowledge about the content of the situation is provided, so that the LLM relies solely on the few-shot examples and the label definitions to perform the task. The prompt is presented in the Appendix \ref{promots} in Figure \ref{fig:prompt1} and \ref{fig:prompt2}.

\subsubsection{Results}

The LLM was applied to all tweets labeled as propagandistic in the HQP dataset \cite{maarouf2023hqp}. In 94 cases, the model did not detect any propagandistic span. Upon manual analysis, we identified that 30 of these cases did exhibit rather clear propagandistic technique or framing. However, without specific contextual knowledge, these cases could often be mistaken for opinion pieces or news. The remaining majority were news reports, discussions, or opinion pieces that did not include explicit propaganda. For the following analysis, we filtered out these cases.\\
The distribution of predicted global labels is summarized in the Appendix \ref{sec:global labels distribution} in Table \ref{tab:global_label_distribution}. The most common labels were \texttt{loaded\_language}, \texttt{doubt}, \texttt{reductio\_ad\_hitlerum}, and \texttt{name\_calling}. Prior work has noted that \texttt{reductio\_ad\_hitlerum} is a frequent technique in Russian propaganda \cite{gherasim2022reductio}. In our setting, this label appears alongside similar categories such as \texttt{loaded\_language} and \texttt{name\_calling}, suggesting empirical overlaps in how these techniques are used. Next, we examined the number of detected propaganda spans per tweet (the distribution is illustrated in Table~\ref {tab:span_distribution}). Our empirical results suggest that a majority of the propagandistic tweets contain multiple propagandistic segments. Relying solely on assigning a global label as has been focused by previous work, may therefore lead to a loss of important details, indicating that future work should maintain the extraction of segments and their local labels as primary target. 
 
\begin{table}[ht]
  \centering
    \small
  \begin{tabular}{lccccc}
    \hline
           \textbf{spans} & \textbf{1} & \textbf{2} & \textbf{3} & \textbf{4} & \textbf{5+} \\
    \hline
    \textbf{count} & 289  & 1,119 & 1,663 & 1,002 & 367 \\
    \hline
  \end{tabular}
    \caption{Distribution of detected Propaganda Spans.}
  \label{tab:span_distribution}
\end{table}

Focusing on tweets with at least three extracted propaganda spans, which is 3,032 cases, we observed that in 76.65\% of these instances, the local label assigned to the first extracted span matched the global label for the entire tweet. This suggests a strong tendency for the most impactful propagandistic content to appear at the beginning of tweets. Furthermore, about 30\% of cases with at least three extractions, exhibited a majority of local labels. In 83.55\% of these cases, this majority local label also aligned with the global label. Thus, we observe that the dominant propaganda technique can be inferred when a majority of extracted local labels is available.

\subsubsection{Ablation}

To assess the robustness of our approach, we conducted several ablation studies. In the first analysis, we compared tweet annotations generated from normalized text (i.e., text with usernames, links, and similar elements removed) against those from non-normalized tweets. To statistically evaluate the differences between these paired categorical observations, we employed the Stuart-Maxwell (marginal homogeneity) test. Under the null hypothesis $H_0$, that the proportion for each predicted global label in the normalized variation is equal to that of the original tweet text. The Stuart-Maxwell test yields a test statistic of 15.32 with 16 degrees of freedom and a $p$-value of 0.5014. Consequently, we conclude that there is no significant difference between the annotated global labels obtained from normalized versus non-normalized text. \\
Next, we evaluated the stability of the LLM’s outputs by repeating the experiment $k=5$ times. Initially, under standard conditions with static few-shot examples, consistent task descriptions, and guided decoding, our approach yielded stable results for the extracted spans, assigned local labels, and the global label in $5/5$ cases. To further challenge the model's robustness, we introduce maximum randomness by shuffling the order of the few-shot examples and the label definitions in the prompt for each data point. We noted the agreement across five runs, randomized for each data point in each run (Table \ref{tab:stability_metrics}). These results indicate that even under maximum prompt randomness, our approach remains quite robust. Nonetheless, variations in the ordering of few-shot examples and label definitions have a marginal effect, particularly on local label predictions, whereas the extracted spans and global label predictions remain more stable. This observation reinforces our initial finding that certain extracted spans may correspond to multiple appropriate labels while still being associated with a consistent global label. 
\begin{table}[ht]
  \centering
    \small
  \begin{tabular}{lccc}
    \hline
    \textbf{Aggreement} & \textbf{$\geq$$3/5$} & \textbf{$\geq$$4/5$} & \textbf{ $5/5$} \\
    \hline

    Local Label    & 100.00\%                    & 95.46\%                   & 81.48\%                     \\
    Extract. Spans & 100.00\%                    & 97.74\%                   & 89.86\%                     \\
    Global Label   & 100.00\%                    & 98.58\%                   & 94.17\%                     \\
    \hline
  \end{tabular}
    \caption{Agreement across $5$ runs with randomization.}
  \label{tab:stability_metrics}
\end{table}

\subsection{Study 2: Human Annotation }
\label{human annotation study 2}
In this second human annotation study, we aim to assess whether integrating LLM-generated annotations with human verification improves annotation consistency and efficiency. Unlike the first study, where annotators assigned coarse- and fine-grained labels without assistance, this study provides them with LLM generated pre-annotations as optional suggestions. Annotators are presented with the original normalized tweet, the extracted spans, and corresponding labels, but they do not modify or verify individual spans. Instead, they select the most appropriate coarse-grained category and fine-grained technique for the entire tweet from a predefined set of options. The predicted global label of the LLM remains hidden while annotating, ensuring that human decisions are less biased and independent of the model's final classification. To minimize potential bias from task familiarity, we exclude the most experienced annotator and swap them with an annotator who has not participated in the first study. This approach is intended to introduce a regularization effect and ensure a more balanced evaluation. 

\subsubsection{Setup}
The annotation process in this study followed the same structured approach as described in the setup in Section \ref{setup1}. However, instead of selecting tweets randomly, we employed a stratified sampling approach based on the global labels predicted by the LLM. Since the distribution of propaganda techniques in real-world data is often imbalanced, random sampling could result in over-representation of some categories and under-representation of others. To ensure that each global label was sufficiently covered, we stratified the sample according to the LLMs predicted global propaganda labels. Most global labels predicted by the LLM appeared frequently in the dataset, allowing for an even allocation across categories. However, techniques such as bandwagon and repetition were considerably less prevalent in the full dataset of 4,534 propagandistic tweets, occurring only 8 times and 6 times, respectively. Based on that, all occurrences of these global labels were included in the sample to ensure that they were adequately represented in the analysis.

\subsubsection{Results}

In the second human annotation study, annotators were provided with LLM-generated pre-annotations that include extracted propagandistic spans along with corresponding local fine-grained labels. However, the predicted global label by the LLM was not shown to them, and annotators remained fully responsible for independently selecting the global coarse- and fine-grained label for each tweet.
Compared to Study 1, this approach led to notable improvements in IAA as well as annotation efficiency.

\begin{figure}
    \centering
    \includegraphics[width=1\linewidth]{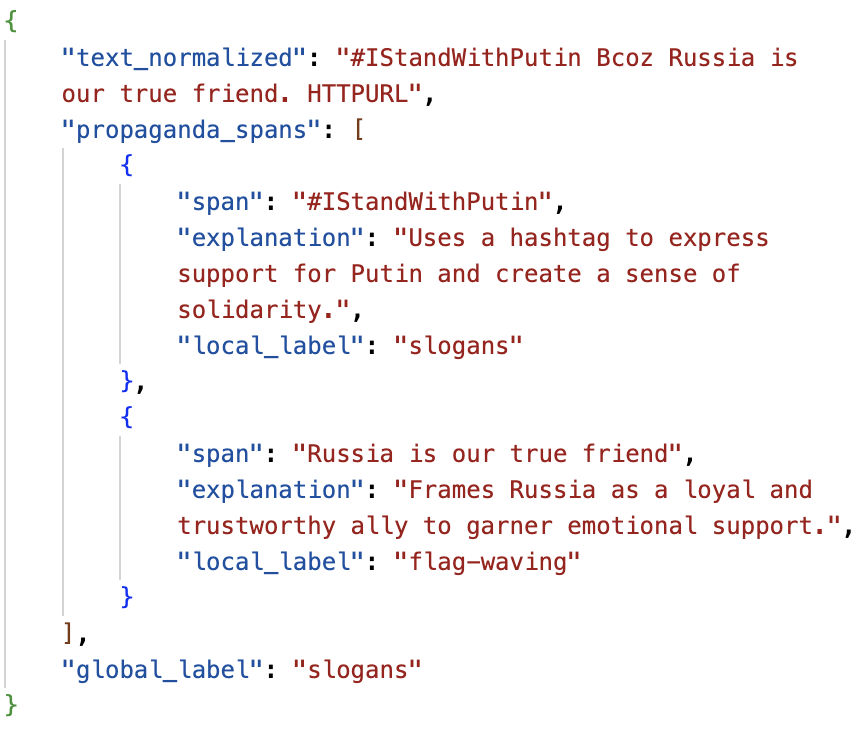}
    \caption{Example of LLM-assisted annotation, showing accurate span extraction, fine-grained label assignment, and coherent explanations. This case achieved full $3/3$ IAA.}
    \label{fig:exampleSlogans}
\end{figure}

\begin{table}[ht]
  \centering
    \small
  \begin{tabular}{lcc}
    \hline

    \hline
    \textbf{Metric}          & \textbf{Coarse} & \textbf{Fine} \\
    \hline
    Raw Agreement $2/3$      &  0.9746       & 0.9014 \\
    Raw Agreement $3/3$      & 0.6225        & 0.4789 \\
    Krippendorff's Alpha     & 0.6059        & 0.5941 \\
    \hline
  \end{tabular}
    \caption{Inter-Annotator Agreement Metrics for Coarse- and Fine-Grained Propaganda Annotations in Round 2.}
  \label{tab:agreement_metrics}
\end{table}
As shown in Table \ref{tab:agreement_metrics}, the raw agreement for coarse-grained labels increased from 88.45\% ($2/3$ majority) and 27.89\% (full consensus) in Study 1 to 97.46\% ($2/3$ majority) and 62.25\% (full consensus) in Study 2. For fine-grained labels, raw agreement improved from 47.61\% $2/3$ and 7.61\% $3/3$ in Study 1 to 90.14\% $2/3$ and 47.89\% in Study 2, respectively. Correspondingly, the Krippendorff's Alpha increased from 0.2065 (coarse) and 0.1233 (fine) in Study 1, to 0.6059 (coarse) and 0.5941 (fine) in Study 2. A detailed examination of fine-grained agreement rates conditioned on the level of consensus in the coarse labels in Table \ref{tab:agreement_2} further confirms these improvements. In Study 2, these rates improved to 80\% for tweets with a (2/3) coarse consensus, and for tweets with full coarse consensus, the (2/3) fine-grained agreement increased to 99.55\%, with full (3/3) fine-grained agreement at 76.02\%. An illustrative example of the effectiveness of LLM-assisted annotation is shown in Figure \ref{fig:exampleSlogans}. In this instance, the LLM successfully identified key propagandistic spans, assigned appropriate fine-grained labels, and provided coherent explanations that aligned well with human interpretations. In this case, the hashtag "\#IStandWithPutin" was labeled as slogans, reinforcing ideological solidarity, while "Russia is our true friend" was classified as flag-waving, portraying Russia as a trustworthy ally. The explanations clearly justify the propagandistic nature of each span, and the global label ("slogans") is particularly suitable, as slogans, especially when used as hashtags, are concise and easily shareable, amplifying their spread on social media and reinforcing group identity more effectively than descriptive statements. This annotation achieved full $3/3$ IAA, confirming its reliability.\\
\begin{table}[ht]
  \centering
    \small
  \begin{tabular}{lcc}
    \hline
    \textbf{Subset}          & \textbf{$2/3$ Fine} & \textbf{$3/3$ Fine} \\
    \hline
    $2/3$ Coarse               &  0.8000  & 0.0160 \\ 
    $3/3$ Coarse               &  0.9955 & 0.7602\\
    \hline
  \end{tabular}
    \caption{Fine-grained agreement rates conditioned on prior majority $2/3$ or full $3/3$ agreement on coarse labels in Round 2.}
  \label{tab:agreement_2}
\end{table}

Additionally, Cohen’s Kappa was calculated to measure agreement between human majority-vote labels and LLM-generated global labels. If no 2/3 majority was reached, a random LLM prediction was used as the human label. The resulting Cohen’s Kappa score of 0.8438 indicates strong agreement between human annotations and LLM-generated global labels. Also, the average annotation time per tweet is reduced from 151.70 seconds in Study 1 to 41.14 seconds in Study 2. In summary, the integration of LLM-generated pre-annotations with human verification in Study 2 resulted in higher IAA and reduces annotation time relative to the fully manual approach in Study 1, indicating an overall improvement in reliability, efficiency and scalability.

\subsection{Knowledge Distillation}
\label{knowledge distillation}

Based on our findings, we next aim to scale structured propaganda annotation and enable efficient inference in resource‐constrained environments by fine‐tuning a collection of SLMs on LLM‐generated supervision. In this knowledge‐distillation‐inspired setup, the 70B model as described in Section \ref{few-shot llm annotation} serves as the \textit{teacher}, providing structured propaganda annotations for every data point. We train four \textit{student} models, two LLaMA3‐based variants (3B and 8B parameters) denoted as \textit{L}, and two Qwen2.5 variants (3B and 7B parameters) denoted as \textit{Q}. To minimize memory usage and accelerate training, we employ parameter‐efficient fine‐tuning (PEFT), combined with 4‐bit quantization. We employ a standard sequence‐to‐sequence cross‐entropy loss, without additional regularization terms or explicit teacher‐student logit matching, to generate the structured responses. We utilize a stratified 80/20 split and learn on the train split for three epochs. 

\subsubsection{Results}
We report six evaluation metrics on the unseen test set as reported in Table \ref{tab:span_prediction_f1_scores}. Here, \textbf{G} denotes the macro- and micro-averaged global F1 scores over the test set. \textbf{Span}$_e$ describes the F1 for exact span detection, while \textbf{Span}$_f$ specifies the fuzzy-span F1 with a strict $0.8$ similarity threshold to account for minor variations following the notion of partial matches as introduced by \cite{hasanain2023large}. Similarly, \textbf{Local}$_e$ requires both exact span text and correct local label classification, while \textbf{Local}$_f$ combines fuzzy span matching with correct local label assignment.

\begin{table}[ht]
  \centering
  \resizebox{\linewidth}{!}{%
    \begin{tabular}{lcccccc}
      \hline
      \textbf{Model} & \textbf{G$_{macro}$} & \textbf{G$_{micro}$} & \textbf{Span}$_e$ & \textbf{Span}$_f$ & \textbf{Local}$_e$ & \textbf{Local}$_f$ \\
      \hline
      \textit{L}$_3$$_b$ & 0.49 & 0.36 & 0.40 & 0.60 & 0.22 & 0.32 \\
      \textit{L}$_8$$_b$ & 0.58 & 0.47 & 0.47 & 0.67 & 0.29 & 0.40 \\
      \textit{Q}$_3$$_b$ & 0.48 & 0.34 & 0.40 & 0.61 & 0.21 & 0.31 \\
      \textit{Q}$_7$$_b$ & 0.51 & 0.34 & 0.45 & 0.66 & 0.25 & 0.36 \\
      \hline
    \end{tabular}
  }
  \caption{Student Model Evaluation Results.}
  \label{tab:span_prediction_f1_scores}
\end{table}

All four student models achieve reasonable performance on each metric. Larger models show modest gains, and \textit{L} and \textit{Q} variants of the same size perform similarly. Global‐label prediction across 14 propaganda categories \cite{martino2020semeval} yields acceptable F1 scores, suggesting that choosing a global label is relatively straightforward. Span detection also works well under both exact‐match and fuzzy‐match criteria. By contrast, assigning local labels remains difficult. Models reliably find propaganda spans but are less certain which specific technique to annotate. We hypothesize that this stems from two key factors: (1) the limited volume of training data available for fine-grained local label predictions, and (2) the inherent ambiguity due to overlap in the definitions of certain propaganda techniques, while the general notion of a propaganda span seems to be more solid.

\section{Discussion and Conclusion}
In this paper, we introduced an LLM-assisted annotation framework that combines automated extraction of propaganda spans with human verification.
Our experiments demonstrate that integrating LLM-assisted pre-annotation with human verification significantly improves the consistency and efficiency of propaganda detection. In Study 1, manual fine-grained labeling suffered from low inter-annotator agreement and long annotation times. Study 2, which incorporated LLM-generated pre-annotations based on extracted propaganda spans, yielded higher agreement metrics and reduced annotation time, although part of the efficiency gain may stem from annotators' familiarity with the task. Notably, our results suggest that a single global label is sometimes insufficient to capture the complexity of propagandistic content, as our analysis shows most tweets include more than one extracted propaganda span. This granular perspective may offer better insights than traditional sequence-level classification, and it is more scalable across different text lengths. These findings, in line with emerging trends such as those highlighted in SemEval-2023 Task 3 \cite{piskorski-etal-2023-semeval}, indicate that future work should consider reformulating the problem to emphasize alternative propaganda detection strategies. Exploring multi-label and hierarchical annotation strategies may better accommodate the overlapping nature of propaganda techniques. Finally, integrating richer contextual information and real-time fact-checking modules could further refine detection performance \cite{sahitaj2025AutomatedFactCheckingRealWorld}. We also advocate for iterative human-in-the-loop systems that continuously update few-shot examples and label definitions to minimize bias and enhance model robustness.

\section*{Limitations}
While promising, our approach has several limitations. First, our study is confined to English tweets related to Russian propaganda which may limit its applicability to other languages or domains. Second, the reliance on a single global label despite the local span-based analysis might oversimplify instances where multiple propaganda techniques coexist. Third, some improvements in annotation efficiency could be attributed to annotator learning effects rather than solely to the LLM-assisted pre-annotation. Fourth, the quality of LLM-generated pre-annotations depends on the few-shot examples and definitions provided which could introduce bias or inconsistencies. Following work should involve a larger and more diverse pool of annotators to further validate and refine the framework. In addition, self-collected data from various propaganda settings encompassing multiple languages and platforms would offer a broader evaluation and help mitigate potential biases inherent in the current dataset. Another limitation concerns our distillation setup. Biases present in the 70B teacher model due to its pretraining may be propagated to the student models. Since the student models are trained solely on model-generated supervision any ideological or geopolitical bias in the teacher can persist without correction. While the use of open-source models improves transparency and auditability it does not inherently prevent bias propagation. Future work should systematically investigate inherited bias in open-source propaganda detection pipelines.

\section*{Ethical and societal implications}

The integration of LLM-assisted annotation in propaganda detection raises ethical concerns regarding bias, automation dependency, misuse, and public trust. While improving annotation efficiency, LLM-generated labels may introduce systematic biases, reflecting dominant narratives in their training data. This can influence human annotators' decisions, leading to reinforced biases instead of neutral classifications. Another risk is automation bias, where annotators overly rely on LLM suggestions and reduce their critical thinking ability. Furthermore, such models could be exploited for counter propaganda, with governments or other actors potentially using them to suppress dissenting voices and shape public discourse to their advantage. Faulty or overly simplistic propaganda detection may inadvertently weaken trust in media and public institutions, undermining the democratic ideals promoted by SDG 16. Therefore, it is imperative that the development and deployment of these systems remain transparent, incorporate rigorous bias audits, and maintain robust human oversight to ensure that they support democratic discourse rather than restrict it.

\section*{Acknowledgments} 
This research is funded by the Federal Ministry of Research, Technology and Space (BMFTR, reference: 03RU2U151C) in the scope of the research project news-polygraph.
  
\bibliography{source}

\appendix

\section{Appendix}
\label{sec:appendix}

\subsection{Fine-grained labels}
\label{sec:fine-grained labels}
The definitions of the propaganda techniques presented here are based on the 14 categories introduced by \citet{martino2020semeval}, which refined an earlier set of 18 techniques proposed in \citet{martino2019fine}. These 14 categories have also been utilized in later works, such as \citet{sprenkamp2023large} and \citet{abdullah2022detecting}, to analyze and label propaganda techniques in text.

\begin{itemize}
    \item[a) ] \textbf{Loaded language} involves the use of words or phrases with either strong positive or negative emotional connotations, to shape audience perceptions and influence their opinions.

    \item[b) ] \textbf{Name calling, labeling} involves assigning a specific label to a target, intended to evoke either positive or negative emotions in the audience, such as fear, hatred, admiration, or praise.

    \item[c) ]\textbf{Repetition} is the continuous repetition of a message or idea to increase its acceptance by the audience over time. 

    \item[d) ]\textbf{Exaggeration or minimization} involves portraying something in an overstated manner to amplify its significance or downplaying its importance to make it appear less impactful than it truly is. 

    \item[e) ]\textbf{Doubt} involves raising uncertainty or questioning the credibility of an individual, group, or entity to undermine trust. 

    \item[f) ]\textbf{Appeal to fear/prejudice} aims to built support for an idea by evoking anxiety, fear, or panic in the audience, often directed at an alternative or based on existing biases. 

    \item[g) ]\textbf{Flag-waving} involves appealing to strong feelings of national or group identity, such as those tied to race, gender, or political affiliation, to justify or promote an action, idea, or individual as representative of the entire group. 

    \item[h) ]\textbf{Causal oversimplification} involves attributing an issue to a single cause while disregarding its complexity or the presence of multiple contributing factors. This may also include assigning blame to an individual or group without adequately exploring the complexity of the issue. 
    
    \item[i) ]\textbf{Slogans} are concise and striking phrases that often incorporate labeling or stereotyping, serving as emotional or cognitive appeals to influence beliefs or perceptions. 

    \item[j) ]\textbf{Appeal to authority} involves asserting that a claim is true solely based on the support of an authority or expert, without providing additional evidence. This can also include cases where the referenced individual lacks genuine expertise but is still presented as authoritative.

    \item[k) ]\textbf{Black-and-white fallacy} involves presenting two opposing options as the only possible choices, disregarding the existence of other alternatives. In its extreme form, referred to as dictatorship, the audience is explicitly directed toward a specific action, effectively eliminating all other options. 

    \item[l) ]\textbf{Thought-terminating cliches} are short, generic phrases designed to suppress critical thinking and meaningful discussion, often by providing oversimplified answers to complex issues or diverting attention from deeper exploration of a topic. 

    \item[m) ]\textbf{Whataboutism, straw man, red herring} combines three distinct techniques, which are frequently grouped together due to their relatively rare individual usage. \textit{Whataboutism} undermines an opponents argument by accusing them of hypocrisy without addressing their claims directly. \textit{Straw man} misrepresents or distorts an opponents position by substituting it with a weaker or exaggerated version that is easier to refute. \textit{Red herring} diverts attention from the main argument by introducing irrelevant information or topics.

    \item[o) ]\textbf{Bandwagon, reductio ad hitlerum} combines two techniques often discussed together due to their similar persuasive nature. \textit{Bandwagon} attempts to convince the audience to adopt an idea or action by emphasizing that "everyone else is doing it". \textit{Reductio ad hitlerum} seeks to discredit an idea or action by associating it with groups or individuals disliked or despised by the audience. 
\end{itemize}

\subsection{Global Labels Distribution}
\label{sec:global labels distribution}
Table~\ref{tab:global_label_distribution} provides an overview of the distribution of global propaganda labels predicted by the model across the dataset. As shown, the most frequently occurring techniques include \texttt{loaded\_language}, \texttt{doubt}, \texttt{reductio\_ad\_hitlerum}, and \texttt{name\_calling}.
\begin{table}[ht]
  \centering
  \caption{Distribution of predicted Global Labels}
  \label{tab:global_label_distribution}
  \begin{tabular}{lc}
    \hline
    \textbf{Global Label} & \textbf{Count} \\
    \hline
    loaded\_language                & 1384 \\
    doubt                           & 647 \\
    reductio\_ad\_hitlerum           & 641 \\
    name\_calling                   & 519 \\
    whataboutism                   & 333 \\
    appeal\_to\_fear\_prejudice     & 250 \\
    causal\_oversimplification       & 160 \\
    exaggeration                   & 150 \\
    flag-waving                    & 122 \\
    appeal\_to\_authority          & 106 \\
    straw\_man                     & 54 \\
    red\_herring                   & 54 \\
    thought-terminating\_cliches   & 35 \\
    slogans                        & 29 \\
    black-and-white\_fallacy       & 25 \\
    repetition                     & 17 \\
    bandwagon                      & 8 \\
    \hline
  \end{tabular}
\end{table}

\subsection{Examples}
\label{sec:examples}

\begin{figure}[ht]
    \centering
    \includegraphics[width=1\linewidth]{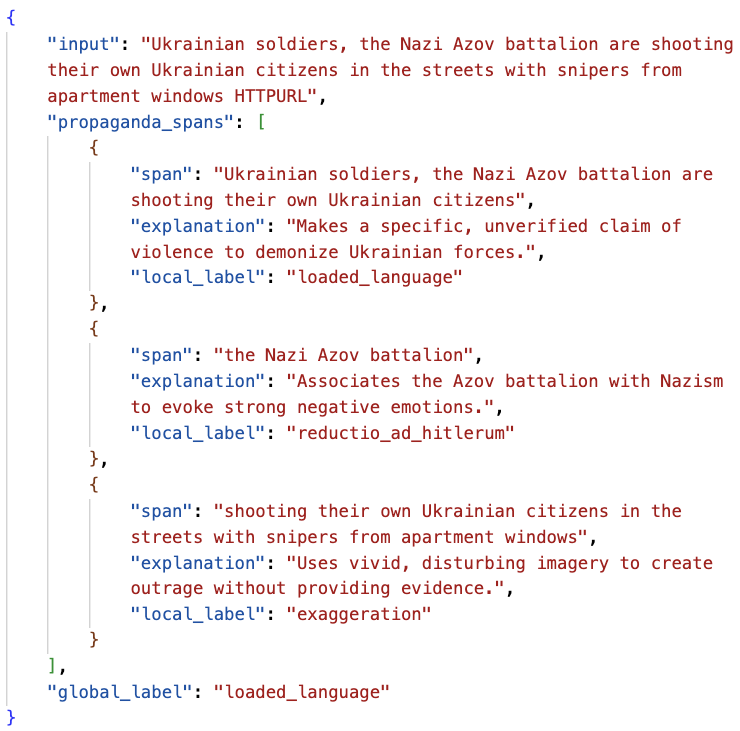}
    \caption{The example highlights how fact-checking is essential, as the detected spans include unverifiable claims of violence ('loaded language'), historical misrepresentation ('reductio ad hitlerum'), and exaggerated imagery ('exaggeration'). Without verification, such statements can contribute to misinformation and manipulation of public perception.}
    \label{fig:factchecking}
\end{figure}

\begin{figure}[ht]
    \centering
    \includegraphics[width=1\linewidth]{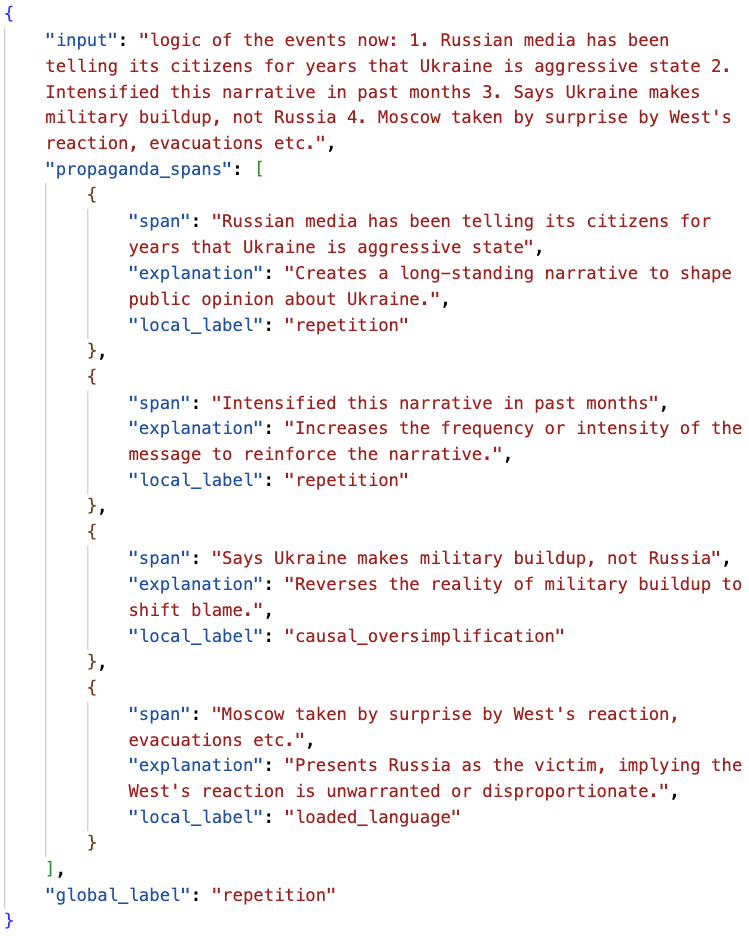}
    \caption{An example initially annotated as propaganda and weak-labeled as 'slogans' in the HQP dataset \cite{maarouf2023hqp}. During qualitative analysis, we found that this example does not necessarily exhibit clear propagandistic intent.}
    \label{fig:apx:exampleno_propaganda}
\end{figure}

In the HQP dataset \cite{maarouf2023hqp}, weak labeling was used to classify certain propaganda techniques. The HQP dataset \cite{maarouf2023hqp} initially weak-labeled this tweet in Figure \ref{fig:factchecking}) as \textit{slogans}. However, a more detailed analysis of the text spans reveals the presence of multiple propaganda techniques, including \textit{loaded language}, \textit{exaggeration}, and \textit{reductio ad Hitlerum}.\\
In the future, by integrating fact-checking into propaganda detection, we can complement existing labeling approaches and assess whether the claims being made have a factual basis. This is important because propaganda often spreads through misinformation, and weak labels alone do not verify truthfulness. Fact-checking strengthens the detection process by distinguishing between persuasive rhetoric and outright disinformation, making it a necessary component for a more precise and reliable analysis of propaganda content. \cite{sahitaj2025AutomatedFactCheckingRealWorld}\\
Another instance of weak labeling challenges in propaganda detection is presented in Figure \ref{fig:apx:exampleno_propaganda}. This example was initially annotated as propaganda in a binary setting by human annotators in the HQP dataset \cite{maarouf2023hqp}. In a later refinement, it was weak-labeled as slogans, reinforcing the classification as propagandistic content. However, during our qualitative analysis, we identified this as a case where the original annotation might not be justified.

\begin{figure}[ht]
    \centering
    \includegraphics[width=1\linewidth]{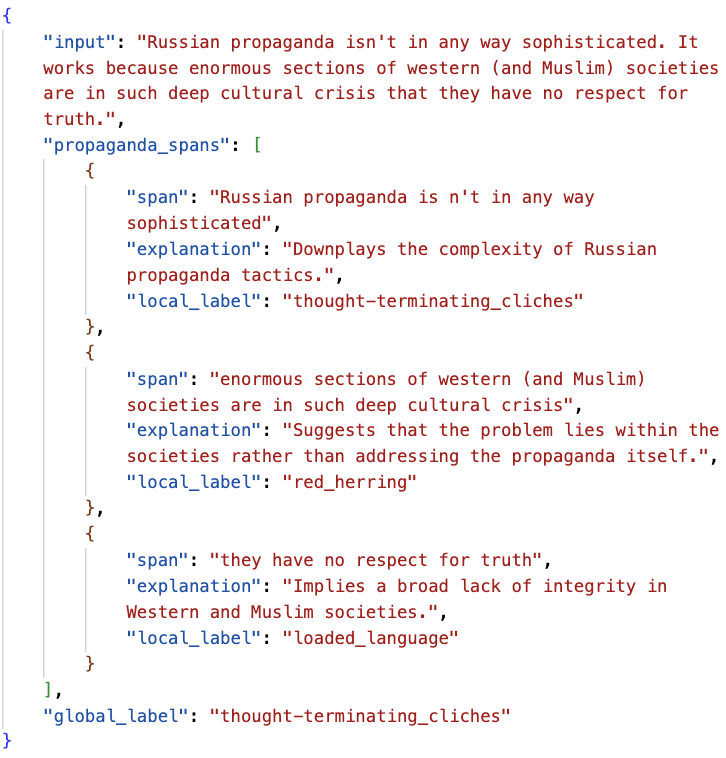}
    \caption{Initially annotated as propaganda and 'loaded language' in the HQP dataset \cite{maarouf2023hqp} and later relabeled using an LLM. While it contains anti-Western criticism, such arguments can exist in normal discourse as well and are not a clear indicator of propaganda.}
    \label{fig:anti_western_criticisml}
\end{figure}

For our analysis, we specifically examined examples that were initially labeled as propaganda before receiving weak labels. This example was among them, but upon closer inspection, we do not find clear propagandistic intent. Instead, the text appears to be an analytical reflection or an ironic commentary on an existing narrative. This case highlights the difficulty of distinguishing between genuine propaganda and discourse that critically engages with a narrative. Automated or weak-labeling approaches may misclassify content that shares linguistic patterns with propaganda but serves a different communicative function. These findings emphasize the need for more nuanced annotation approaches that incorporate contextual understanding, ensuring that content is not misclassified based solely on surface-level textual features.
Another example illustrating the complexity of propaganda detection is shown in Figure \ref{fig:anti_western_criticisml}. This statement was initially annotated as propaganda in the HQP dataset \cite{maarouf2023hqp} and subsequently relabeled using an LLM-based approach. The LLM did not perform binary classification but instead assigned fine-grained rhetorical labels, identifying thought-terminating clichés, red herring, and loaded language. However, during a qualitative review, we noticed that this example raises important questions about what should and should not be considered propaganda.
This case is particularly interesting because, while the statement contains anti-Western criticism, which can be used in propaganda, it can also appear in normal discussions and political discourse. It does not necessarily display clear manipulative intent, even though it employs rhetorical techniques often associated with propaganda. The key challenge here is that rhetorical strategies alone do not automatically make a statement propagandistic. Context and intent matter. During our qualitative investigation of tweets, we found this to be a borderline case where one could argue both for and against labeling it as propaganda. On the one hand, its framing could serve as a tool for propaganda by reinforcing narratives about Western decline. On the other hand, such critiques exist independently of propaganda efforts. This example is valuable because it demonstrates that the LLM correctly assigned rhetorical strategies without overgeneralizing the statement as propaganda, highlighting the difficulty of drawing a clear boundary between manipulative content and critical discussion.

\subsection{Prompts}
\label{promots}

The prompt establishes a structured framework for LLM-assisted annotation in propaganda detection, defining a systematic approach for identifying, explaining, and categorizing propagandistic content. As shown in Figures \ref{fig:prompt1} and \ref{fig:prompt2}, the assistant is designed to extract specific spans indicative of propaganda, provide justifications based on predefined classification criteria, and assign both fine-grained local labels and an overarching global label. The framework (Figure \ref{fig:prompt1}) first guides the assistant to detect key propaganda spans, classify them based on a predefined set of propaganda techniques, and explain why each span should be considered propaganda.

\begin{figure}[ht]
  \centering
\begin{tcolorbox}[colback=white!10!white, colframe=white!50!black, title=Prompt]
\scriptsize
\begin{verbatim}
SYSTEM:
You are an intelligent annotation assistant specializing in
detecting propaganda. Your task is to analyze, explain, and 
pre-annotate the presented text based on a set of potential
propaganda classifications. You MUST return the output in
valid JSON following the defined schema.

**Setting**: Detection of propaganda that is against the 
main opposition (i.e., Ukraine), against other oppositions 
(e.g., Western countries), or in favour of the Russian 
government. 

1. **Identify specific words or text spans 
     that indicate propaganda.**:

2. **Explain for each extracted span why it 
     should be considered propaganda.**:

3. **For each span, determine the dominant 
    propaganda technique from the following list**:
    - Loaded language: ...
    - Name calling: ...
    - Appeal to fear/prejudice: ...
    - Flag-waving: ...
    - Slogans: ...
    - Repetition: ...
    - Exaggeration/minimization: ...
    - Causal oversimplification: ...
    - Black-and-white fallacy: ...
    - Thought-Terminating Cliches: ...
    - Doubt: ...
    - Appeal to authority: ...
    - Whataboutism: ...
    - Straw man: ...
    - Red herring: ...
    - Bandwagon: ...
    - Reductio ad hitlerum: ...
4. **Finally, assign the global label of the span that 
     is most representative for the full sequence.**

\end{verbatim}
\end{tcolorbox}
\caption{Prompt (Part 1): Initial instructions for the propaganda detection task, including span extraction, explanation, and classification of local and global labels.}
  \label{fig:prompt1}
\end{figure}
The second part (Figure \ref{fig:prompt2}) extends this process by enforcing a structured JSON output format, ensuring consistency across annotations and facilitating integration with human verification workflows. By structuring the annotation process in this way, our approach aims to improve labeling efficiency, reduce inter-annotator variability, and enhance the scalability of propaganda detection in large-scale datasets. The explicit categorization of rhetorical techniques provides a more detailed understanding of how propaganda manifests in text, while the standardized output format ensures that annotations remain interpretable and reproducible.

\begin{figure}[t]
  \centering
\begin{tcolorbox}[colback=white!10!white, colframe=white!50!black]
\scriptsize
\begin{verbatim}
**Output Format:**
Respond in **valid JSON** with the structure:
{
    "$defs": {
        "FineLabelVerdict": {
            "description": "Fine-grained categorization of
            propaganda techniques.",
            "enum": [
                ${LABELS}
            ]
        },
        "PropagandaSpan": {
            "description": "An identified propaganda span 
            within the original text with an explanation.",
            "properties": {
                "span": {
                    "description": "The exact propaganda span
                    extracted from the original text.",
                    "title": "Span",
                    "type": "string"
                },
                "explanation": {
                    "description": "The explanation why this
                    span is considered propaganda.",
                    "title": "Explanation",
                    "type": "string"
                },
                "local_label": {
                    "$ref": "#/$defs/FineLabelVerdict",
                    "description": "The appropriate label 
                    assigned towards the detected label."
                }
            },
            "required": [
                "span",
                "explanation",
                "local_label"
            ]
        },
        "global_label": {
            "$ref": "#/$defs/FineLabelVerdict",
            "description": "The label for the dominant 
            propaganda technique in the statement."
        }
    },
    "description": "Schema for structured LLM output after
    propaganda detection and normalization."
}
USER:
${TWEET}

ASSISTANT:
\end{verbatim}
\end{tcolorbox}
\caption{Prompt (Part 2): JSON output format definition for our propaganda detection task.}
  \label{fig:prompt2}
\end{figure}

\end{document}